\documentclass[11pt,a4paper]{article}
\usepackage[hyperref]{naaclhlt2019}
\usepackage{times}
\usepackage{latexsym}

\usepackage{url}

\usepackage{graphicx}
\usepackage{amsmath}
\usepackage{bm}
\usepackage[tone]{tipa}
\usepackage{cleveref}
\crefname{section}{§}{§§}
\Crefname{section}{Section}{}
\Crefname{figure}{Figure}{}
\Crefname{algorithm}{Algorithm}{}
\Crefname{equation}{Equation}{}
\usepackage{shadowtext}
\usepackage{xspace}
\newcommand{\ReCoRD}{{\fontfamily{fla}\fontseries{m}\selectfont Re}{\fontfamily{fla}\fontseries{b}\selectfont{Co}}{\fontfamily{fla}\fontseries{m}\selectfont RD}\xspace}
\DeclareMathOperator*{\argmax}{arg\,max}
\usepackage{booktabs}
\usepackage{multirow}

\usepackage{array}
\newcolumntype{L}[1]{>{\raggedright\let\newline\\\arraybackslash\hspace{0pt}}m{#1}}
\newcolumntype{C}[1]{>{\centering\let\newline\\\arraybackslash\hspace{0pt}}m{#1}}
\newcolumntype{R}[1]{>{\raggedleft\let\newline\\\arraybackslash\hspace{0pt}}m{#1}}
\usepackage{enumitem}
\usepackage{todonotes} 

\newcommand{\kev}[2][disable]{\todo[color=purple!20,size=\scriptsize,fancyline,caption={},#1]{kev:#2}}
\newcommand{\ben}[2][disable]{\todo[color=red!20,size=\scriptsize,fancyline,caption={},#1]{ben:#2}}

\aclfinalcopy 

\title{{\fontfamily{fla}\fontseries{m}\selectfont Re}{\fontfamily{fla}\fontseries{b}\selectfont \hspace{-0.28ex}\shadowtext{Co}}\hspace{-0.08ex}{\fontfamily{fla}\fontseries{m}\selectfont RD}: 
Bridging the Gap between Human \\ and Machine Commonsense Reading Comprehension}

\author{Sheng Zhang$^\dagger$\thanks{~~Work done when Sheng Zhang was visiting Microsoft.}, Xiaodong Liu$^\ddagger$, Jingjing Liu$^\ddagger$, Jianfeng Gao$^\ddagger$, \\ 
{\bf Kevin Duh$^\dagger$ \and Benjamin Van Durme$^\dagger$} \\
$^\dagger${Johns Hopkins University}\\
$^\ddagger${Microsoft Research}\\
}

\date{}

\begin{document}
\maketitle
\begin{abstract}
We present a large-scale dataset, \ReCoRD,
for machine reading comprehension requiring commonsense reasoning.
Experiments on this dataset demonstrate that the performance of state-of-the-art MRC systems fall far behind human performance.
\ReCoRD represents a challenge for future research to bridge the gap between human and machine commonsense reading comprehension. 
\ReCoRD is available at \url{http://nlp.jhu.edu/record}.
\kev{How about emphasize more the motivation of common sense more in the abstract?}
\end{abstract}

\section{Introduction}
\ben{It is a little weird that RECORD is not spelled out in the abstract, but especially odd that it isn't spelled out in the Introduction.  I would remove the footnote, put that content in the Introduction}

\ben{@kev agree.  ... Human and Machine Commonsense Reading Comprehension}

\ben{Methods in machine reading comprehension (MRC) are driven by the datasets available -- such as curated by \newcite{deepmind-cnn-dailymail}, \newcite{cbt}, \newcite{squad}, \newcite{newsqa}, and \newcite{msmarco} -- where an MRC task is commonly defined as answering a question given some passage. However ...}

Machine reading comprehension (MRC) is a central task in natural language understanding, with techniques lately driven by a surge of large-scale datasets~\cite{deepmind-cnn-dailymail,cbt,squad,newsqa,msmarco}, usually formalized as a task of answering questions given a passage. An increasing number of analyses~\cite{adversarial-squad,squad-v2,how-much-reading} have revealed that a large portion of questions in these datasets can be answered by simply matching the patterns between the question and the answer sentence in the passage.
While systems may match or even outperform humans on these datasets, our intuition suggests that there are at least some instances in human reading  comprehension that require more than what existing challenge tasks are emphasizing.
\ben{This "thus" claim is far too strong.  You haven't cited anything that says humans *don't* rely on simple pattern matching, you just rely on an implicit assumption that 'surely humans must be doing something complicated when they read'.  If a system performs as well as a human on a task, the conclusion shouldn't immediately be that the task is too easy, it should more subtly be that new datasets are then needed to see if the inference mechanisms hold up, where the creation of the datasets can be based based on an explicitly stated intuition that humans \emph{may} rely on more than pattern matching.  It is a hypothesis at this point in the Introduction, that systems doing well on earlier datasets won't also do well on yours.  You expect they will fail, and even design the dataset specifically around their failure cases. }
\ben{I would say: While systems may match or even outperform humans on these datasets, our intuition suggests that there are at least some instances in human reading  comprehension that require more than what existing challenge tasks are stressing.}
One primary type of questions these datasets lack are the ones that require reasoning over common sense or understanding across multiple sentences in the passage~\cite{squad,newsqa}.
\ben{This statement is given without citation: why do you claim that common sense is missing?  Do you provide an analysis later in this paper that supports it?  If so, provide a forward reference.  If you can cite earlier work, do so.  Otherwise, remove or soften this statement, e.g., "We hypothesize that one type of question ...".  And then in next sentence, rather than "To overcome this limitation", which you haven't proven yet actually exists, you would say: "To help evaluate this question, we introduce ..."}
\ben{rather than "most of which require", say "most of which seem to require some aspect of reasoning beyond immediate pattern matching".  The SWAG / BERT case should be fresh in your mind as you write this introduction, and where-ever you are tempted to declare things in absolute terms.   The more you go on the record as THIS DATASET REQUIRES COMMONSENSE then the more you look silly later if someone finds a 'trick' to solve it.  A more honest and safer way to put this is to exactly reference the SWAG/BERT issue at some point in this paper, acknowledging that prior claims to have constructed commonsense datasets have been shown to either be false, or to imply that commonsense reasoning can be equated to large scale language modeling.  You can cite Rachel's Script Induction as Language Modeling paper, JOCI,  and the reporting bias article, perhaps all in a footnote, when commenting that researchers have previously raised concerns about the idea that all of common sense can be derived from corpus co-occurrence statistics.}

\begin{figure}[!t]
\centering
\includegraphics[width=0.48\textwidth]{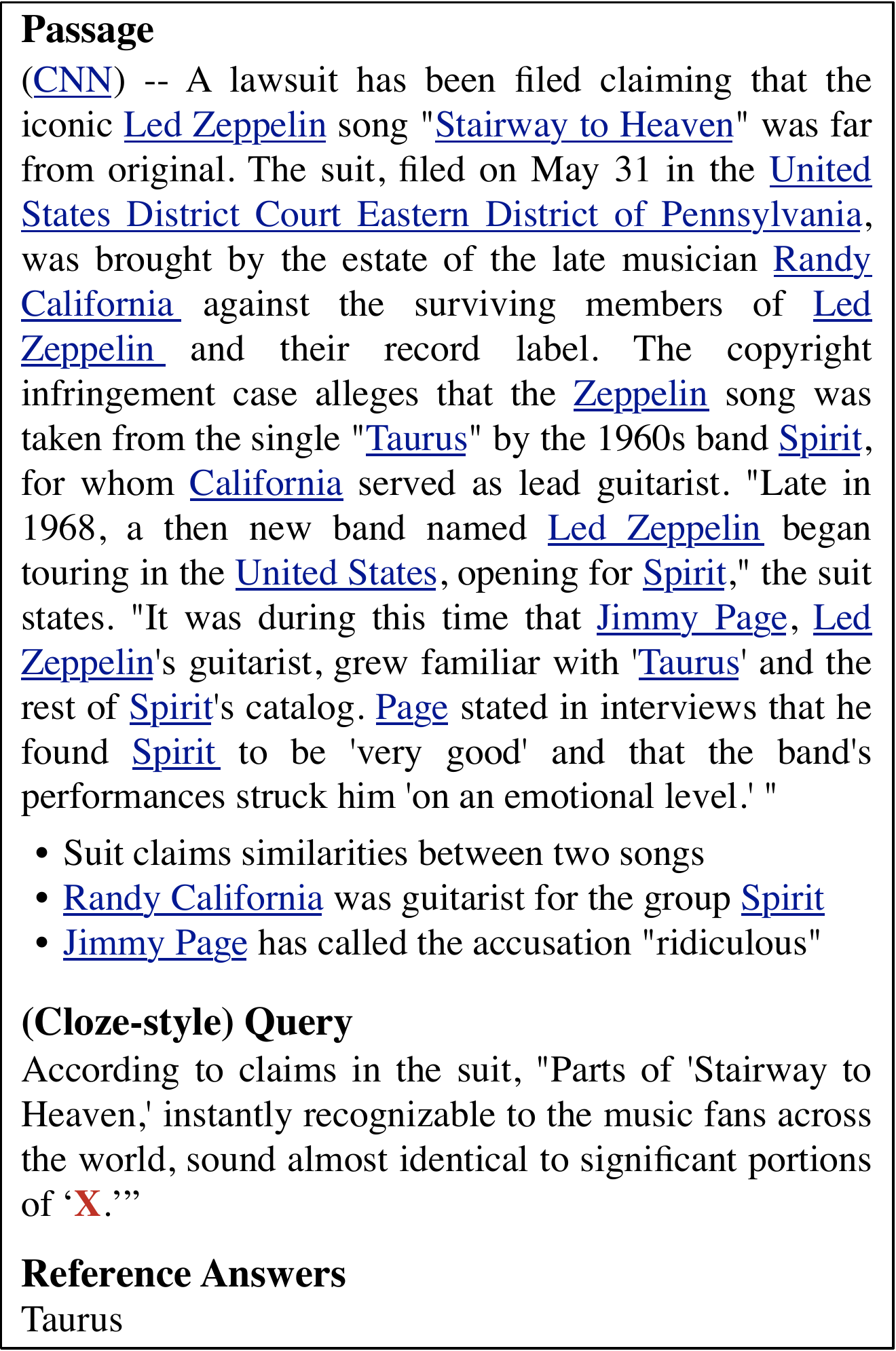}
\caption{An example from \ReCoRD. 
The \textbf{passage} is a snippet from a news article followed by some bullet points which summarize the news event. Named entities highlighted in the passage are possible answers to the query.
The \textbf{query} is a statement that is factually supported by the passage.
    $\mathbf{X}$ in the statement indicates a missing named entity.
    The goal is to find the correct entity in the passage that best fits $\mathbf{X}$.\label{fig:example}}
\end{figure}

To overcome this limitation, we introduce a large-scale dataset for reading comprehension, \ReCoRD (\textipa{["rEk@rd]}), which consists of over 120,000 examples, most of which require deep commonsense reasoning.
\ReCoRD is an acronym for the \textbf{Re}ading \textbf{Co}mprehension with \textbf{Co}mmonsense \textbf{R}easoning \textbf{D}ataset. 

\Cref{fig:example} shows a \ReCoRD example: 
the passage describes a lawsuit claiming that the band  ``\emph{Led Zeppelin}'' had plagiarized the song ``\emph{Taurus}'' to their most iconic song, ``\emph{Stairway to Heaven}''.
The cloze-style query asks what does ``\emph{Stairway to Heaven}'' sound similar to. 
To find the correct answer, we need to understand from the passage that ``\emph{a copyright infringement case alleges that `Stairway to Heaven' was taken from `Taurus'}'', and from the bullet point that ``\emph{these two songs are claimed similar}''. 
Then based on the commonsense knowledge that ``\emph{if two songs are claimed similar, it is likely that (parts of) these songs sound almost identical}'', we can reasonably infer that the answer is ``\emph{Taurus}''.
\kev{This example is good, but you might need to make sure the reader reads the whole passage first or else it may be hard to follow. Maybe add a few more sentences to explain Figure 1 in the paragraph here.} 

Differing from most of the existing MRC datasets, all queries and passages in \ReCoRD are automatically mined from news articles, which maximally reduces the human elicitation bias \cite{reporting-bias,vision-reporting-bias,joci},
and the data collection method we propose is cost-efficient.
\kev{You should have one of these comparison tables that lists multiple MRC datasets and compares different features}
Further analysis shows that
a large portion of \ReCoRD requires commonsense reasoning.

Experiments on \ReCoRD demonstrate that human readers are able to achieve a high performance at 91.69 F1, whereas the state-of-the-art MRC models fall far behind at 46.65 F1.
Thus, \ReCoRD presents a real challenge for future research to bridge the gap between human and machine commonsense reading comprehension.
\ben{this is a bulky URL: I will pay the small fee to register some domain name that is more slick than this}
\ben{about the leaderboard on the website: I think it a little misleading to have Google Brain and IBM Watson, etc. as the names on the leaderboard, if it is really you running their code.  Better would be "JHU (modification of Google Brain system)", "JHU (modification of IBM Watson system)", ... .}

\section{Task Motivation}
\label{sec:task}

\begin{quote}
  {\emph{A program has common sense if it automatically deduces for itself a sufficiently wide class of immediate consequences of anything it is told and what it already knows.} --~\newcite{mccarthy59}}
\end{quote}

\noindent\textbf{Commonsense Reasoning in MRC}
As illustrated by the example in \Cref{fig:example}, the commonsense knowledge ``\emph{if two songs are claimed similar, it is likely that (parts of) these songs sound almost identica}'' is not explicitly described in the passage, but is necessary to acquire in order to generate the answer.
Human is able to infer the answer because the commonsense knowledge is commonly known by nearly all people.
Our goal is to evaluate whether a machine is able to learn such knowledge.
However, since commonsense knowledge is massive and mostly implicit, defining an explicit free-form evaluation is challenging~\cite{wsc}.
Motivated by \newcite{mccarthy59}, we instead evaluate a machine's ability of commonsense reasoning -- a reasoning process requiring commonsense knowledge; that is, if a machine has common sense, it can deduce for itself the likely consequences or details of anything it is told and what it already knows rather than the unlikely ones.
To formalize it in MRC, given a passage $\mathbf{p}$ (i.e., ``\emph{anything it is told}'' and ``\emph{what it already knows}''),
and a set of consequences or details $\mathcal{C}$ which are factually supported by the passage $\mathbf{p}$ with different likelihood,
if a machine $\mathbf{M}$ has common sense, it can choose the most likely consequence or detail $\mathbf{c}^*$ from $\mathcal{C}$, i.e., 
\begin{equation}
\label{eq:csr-in-mrc}
    \mathbf{c}^* = \argmax_{\mathbf{c} \in \mathcal{C}}P(\mathbf{c}\mid\mathbf{p},\mathbf{M}).
\end{equation}

\kev{What are the properties of $o$? What can be a consequence? Be more specific or give examples.}

\noindent\textbf{Task Definition} With the above discussion, we propose a specific task to evaluate a machine's ability of commonsense reasoning in MRC: as shown in \Cref{fig:example}, given a passage $\mathbf{p}$ describing an event, a set of text spans $\mathbf{E}$ marked in $\mathbf{p}$, and a cloze-style query $Q(\mathbf{X})$ with a missing text span indicated by $\mathbf{X}$,
a machine $\mathbf{M}$ is expected to act like human, reading the passage $\mathbf{p}$ and then using its hidden commonsense knowledge to choose a text span $\mathbf{e}\in\mathbf{E}$ that best fits $\mathbf{X}$, i.e., 
\begin{equation}
\label{eq:task}
    \mathbf{e}^* = \argmax_{\mathbf{e} \in \mathbf{E}}P(Q(\mathbf{e})\mid\mathbf{p},\mathbf{M}).
\end{equation}

Once the cloze-style query $Q(\mathbf{X})$ is filled in by a text span $\mathbf{e}$, the resulted statement $Q(\mathbf{e})$ becomes a consequence or detail $\mathbf{c}$ as described in \Cref{eq:csr-in-mrc}, which is factually supported by the passage with certain likelihood.

\kev{There's a disconnect between this paragraph and the previous one. How do you jump from $o$ to Q(e) and the ineqality to argmax? Also, I'm not sure if "cloze" is defined anywhere: you might need a one-sentence explanation in case the reader is not familiar.}

\section{Data Collection}

\begin{figure}[!ht]
\centering
\includegraphics[width=0.45\textwidth]{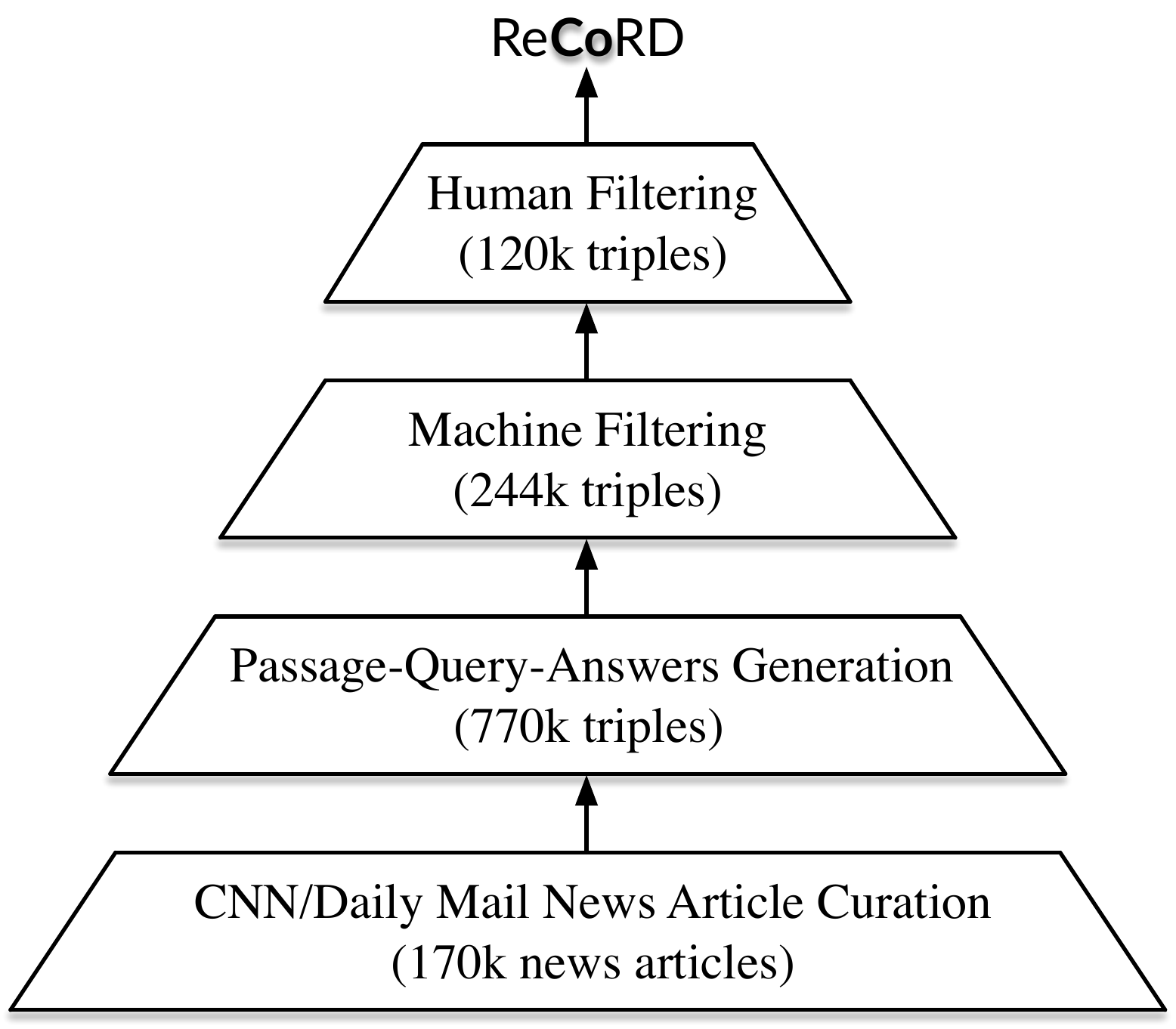}
\caption{The overview of data collection stages.\label{fig:flow-chart}}
\end{figure}

\kev{First add motivation about general philosophy of data collection} 
We describe the framework for automatically generating the dataset, \ReCoRD, for our task defined in \Cref{eq:task},
which consists of passages with text spans marked, cloze-style queries, and reference answers.
We collect \ReCoRD in four stages as shown in Figure~\ref{fig:flow-chart}:
(1) curating CNN/Daily Mail news articles,
(2) generating passage-query-answers triples based on the news articles,
(3) filtering out the queries that can be easily answered by state-of-the-art MRC models,
and (4) filtering out the queries ambiguous to human readers.

\begin{figure*}[!t]
\centering
\includegraphics[width=0.98\textwidth]{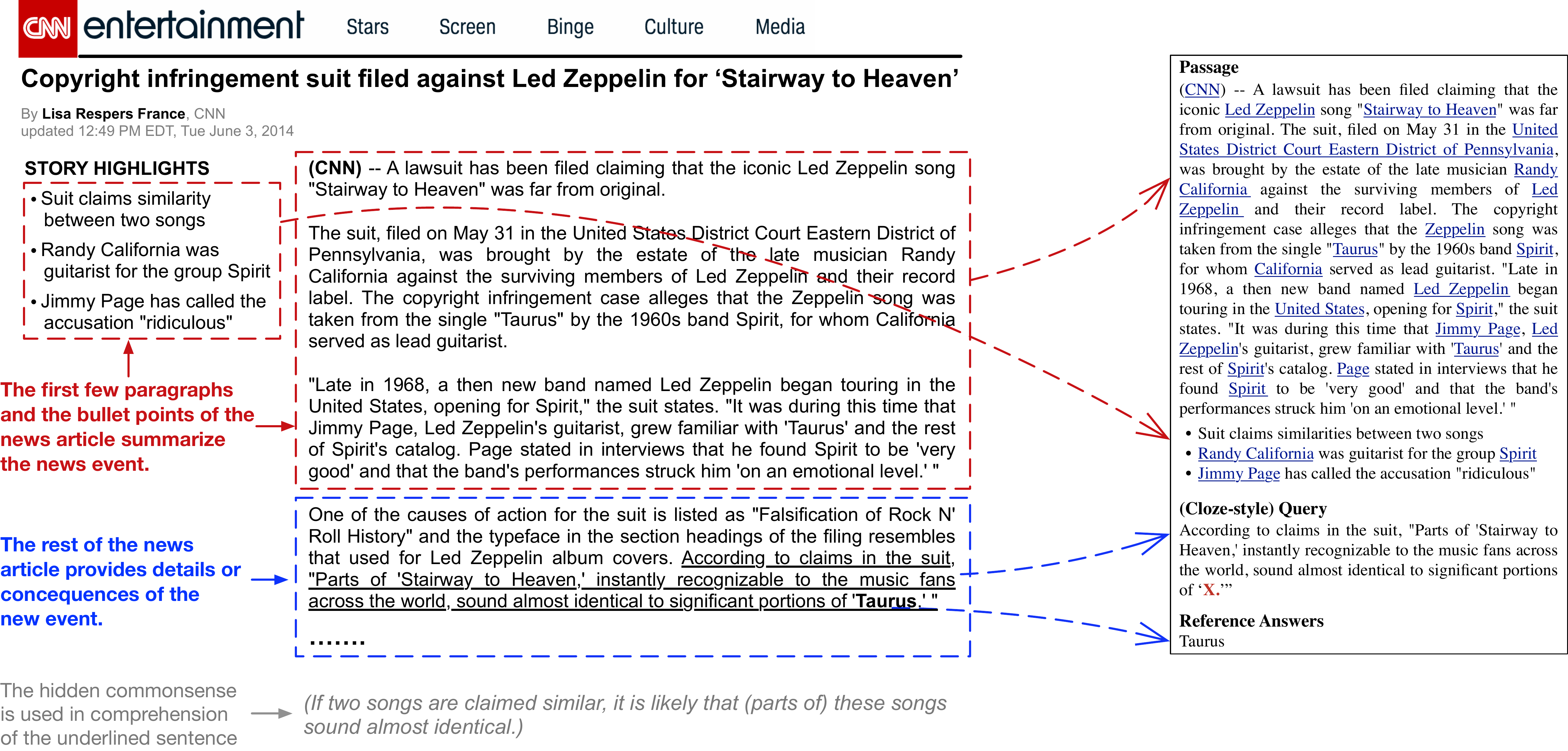}
\caption{Passage-query-answers generation from a CNN news article.\label{fig:example-for-stage2}}
\end{figure*}

\subsection{News Article Curation}
\label{sec:news-curation}
We choose to create \ReCoRD by exploiting news articles, because the structure of news makes it a good source for our task:
normally, the first few paragraphs of a news article summarize the news event, which can be used to generate passages of the task;
and the rest of the news article provides consequences or details of the news event, which can be used to generate queries of the task.
In addition, news providers such as CNN and Daily Mail supplement their articles with a number of bullet points~\cite{Ranknet,story-highlight-generation,deepmind-cnn-dailymail}, which outline the highlights of the news and hence form a supplemental source for generating passages.

We first downloaded CNN and Daily Mail news articles using the script\footnote{\url{https://github.com/deepmind/rc-data}} provided by~\citet{deepmind-cnn-dailymail}, 
and then sampled 148K articles from CNN and Daily Mail.
In these articles, named entities and their coreference information have been annotated by a Google NLP pipeline, and will be used in the second stage of our data collection.
Since these articles can be easily downloaded using the public script, we are concerned about potential cheating if using them as the source for generating the dev./test datasets.
Therefore, we crawled additional 22K news articles from the CNN and Daily Mail websites.
These crawled articles have no overlap with the articles used in~\citet{deepmind-cnn-dailymail}.
We then ran the state-of-the-art named entity recognition model~\cite{elmo} and the end-to-end coreference resolution model~\cite{end-to-end-coref} provided by AllenNLP~\cite{allennlp} to annotate the crawled articles.
Overall, we have collected 170K CNN/Daily Mail news articles with their named entities and coreference information annotated.

\subsection{Passage-Query-Answers Generation}
All passages, queries and answers in \ReCoRD were automatically generated from the curated news articles.
\Cref{fig:example-for-stage2} illustrates the generation process. (1) we split each news article into two parts as described in \Cref{sec:news-curation}:
the first few paragraphs which summarize the news event,
and the rest of the news which provides the details or consequences of the news event.
These two parts make a good source for generating passages and queries of our task respectively.
(2) we enriched the first part of news article with the bullet points provided by the news editors.
The first part of news article, together with the bullet points, is considered as a candidate passage.
To ensure that the candidate passages are informative enough, we required the first part of news article to have at least 100 tokens and contain at least four different entities.
(3) for each candidate passage, the second part of its corresponding news article was split into sentences by Stanford CoreNLP \cite{corenlp}.
Then we selected the sentences that satisfy the following conditions as potential details or consequences of the news event described by the passage: 
\begin{itemize}[itemsep=0pt,topsep=6pt,leftmargin=10pt]
    \item Sentences should have at least 10 tokens, as longer sentences contain more information and thus are more likely to be inferrable details or consequences.
    \item Sentences should not be questions, as we only consider details or consequences of a news event, not questions.
    \item Sentences should not have 3-gram overlap with the corresponding passage, so they are less likely to be paraphrase of sentences in the passage.
    \item Sentences should have at least one named entity, so that we can replace it with $\mathbf{X}$ to generate a cloze-style query.
    \item All named entities in sentences should have precedents in the passage according to coreference, so that the sentences are not too disconnected from the passage, and the correct entity can be found in the passage to fill in $\mathbf{X}$. 
\end{itemize}
Finally, we generated queries by replacing entities in the selected sentences with $\mathbf{X}$.  
We only replaced one entity in the selected sentence each time, and generated one cloze-style query.
Based on coreference, the precedents of the replaced entity in the passage became reference answers to the query. 
The passage-query-answers generation process matched our task definition in \Cref{sec:task}, and therefore created queries that require some aspect of reasoning beyond immediate pattern matching.
In total, we generated 770k (passage, query, answers) triples.

\subsection{Machine Filtering}
As discussed in \citet{adversarial-squad,squad-v2,adversarial-training,how-much-reading},
existing MRC models mostly learn to predict the answer by simply paraphrasing questions into declarative forms, and then matching them with the sentences in the passages.
To overcome this limitation, we filtered out triples whose queries can be easily answered by
the state-of-the-art MRC architecture, Stochastic Answer Networks (SAN) \cite{san}.
We choose SAN because it is competitive on existing MRC datasets, and it has components widely used in many MRC architectures such that low bias was anticipated in the filtering (which is confirmed by evaluation in \Cref{sec:evaluation}). 
We used SAN to perform a five-fold cross validation on all 770k triples.
The SAN models correctly answered 68\% of these triples. 
We excluded those triples, and only kept 244k triples that could not be answered by SAN. 
These triples contain queries which could not be answered by simple paraphrasing, and other types of reasoning such as commonsense reasoning and multi-sentence reasoning are needed.
\kev{Briefly mention why you use SAN, i.e. it's competitive on current benchmarks like SQuAD. Also mention whether this may cause some bias in the filtering, compared to using some other system, and why your methodology is still ok.}

\subsection{Human Filtering}
\label{sec:human-filtering}
Since the first three stages of data collection were fully automated,
the resulted triples could be noisy and ambiguous to human readers.
Therefore, we employed crowdworkers to validate these triples.
We used Amazon Mechanical Turk for validation.
Crowdworkers were required to: 1) have a 95\% HIT acceptance rate, 2) a minimum of 50 HITs, 3) be located in the United States, Canada, or Great Britain, and 4) not be granted the qualification of poor quality (which we will explain later in this section).
Workers were asked to spend at least 30 seconds on each assignment, and paid \$3.6 per hour on average.

\begin{figure}[!ht]
\centering
\includegraphics[width=0.49\textwidth]{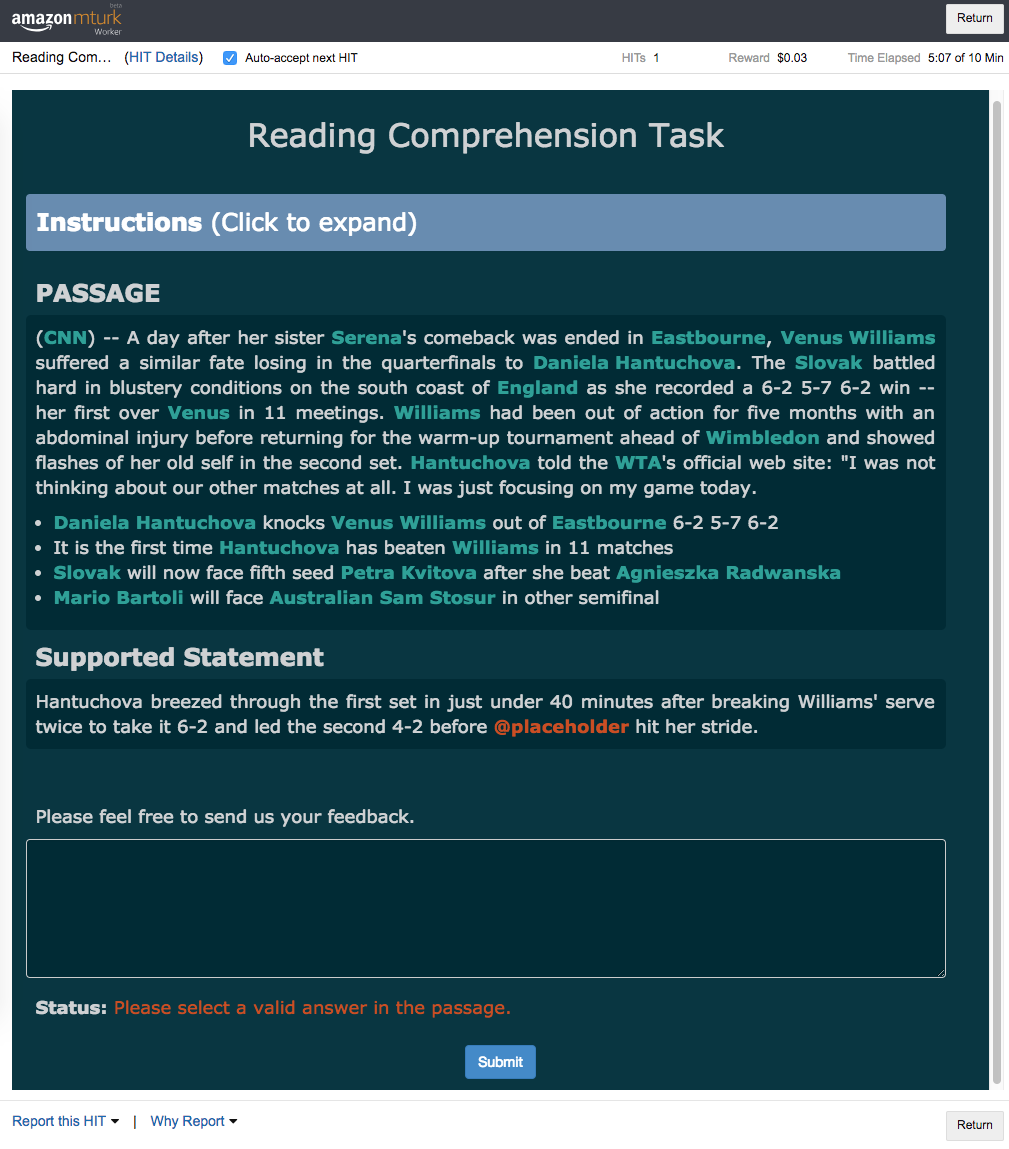}
\caption{The crowdsourcing web interface.\label{fig:hit}}
\end{figure}

\Cref{fig:hit} shows the crowdsourcing web interface.
Each HIT corresponds to a triple in our data collection.
In each HIT assignment, we first showed the expandable instructions for first-time workers, to help them better understand our task (see the \Cref{sec:hit-instructions}).
Then we presented workers with a passage in which the named entities are highlighted and clickable.
After reading the passage, workers were given a supported statement with a placeholder (i.e., a cloze-style query) indicating a missing entity.
Based on their understanding of the events that might be inferred from the passage, workers were asked to find the correct entity in the passage that best fits the placeholder. 
If workers thought the answer is not obvious, they were allowed to guess one, and were required to report that case in the feedback box.
Workers were also encouraged to write other feedback.

To ensure quality and prevent spamming, we used the reference answers in the triples to compute workers' average performance after every 1000 submissions. 
While there might be coreference or named entity recognition errors in the reference answers, 
as reported in \citet{exam-of-cnn-dailymail} (also confirmed by our analysis in \Cref{sec:data-analysis}), they only accounted for a very small portion of all the reference answers. Thus, the reference answers could be used for comparing workers' performance. 
Specifically, if a worker's performance was significantly lower than the average performance of all workers, we blocked the worker by granting the qualification of poor quality.
In practice, workers were able to correctly answer about 50\% of all queries. 
We blocked workers if their average accuracy was lower than 20\%, and then republished their HIT assignments.
Overall, 2,257 crowdworkers have participated in our task, and 51 of them have been granted the qualification of poor quality.

\noindent\textbf{Train\,/\,Dev.\,/\,Test Splits}
Among all the 244k triples collected from the third stage, we first obtained one worker answer for each triple. 
Compared to the reference answers, workers correctly answered queries in 122k triples.
We then selected around 100k correctly-answered triples as the training set,
restricting the origins of these triples to the news articles used in \citet{deepmind-cnn-dailymail}.
As for the development and test sets, we solicited another worker answer to further ensure their quality.
Therefore, each of the rest 22k triples has been validated by two workers.
We only kept 20k triples that were correctly answered by both workers.
The origins of these triples are either articles used in \citet{deepmind-cnn-dailymail} or articles crawled by us (as described in \Cref{sec:news-curation}), with a ratio of 3:7.
Finally, we randomly split the 20k triples into development and test sets, with 10k triples for each set.
\Cref{tab:statistics} summarizes the statistics of our dataset, \ReCoRD.

\begin{table}[!ht]
\small
\begin{tabular}{@{}l|r|r|r|r@{}}
\toprule
                     & \multicolumn{1}{c|}{Train}   & \multicolumn{1}{c|}{Dev.}   & \multicolumn{1}{c|}{Test}   & \multicolumn{1}{c}{Overall} \\ \midrule
queries          & 100,730                      & 10,000                      & 10,000                      & 120,730                     \\
unique passages    & 65,709                       & 7,133                       & 7,279                       & 80,121                      \\ \midrule
passage vocab.       & 352,491 & 93,171 & 94,386 & 395,356 \\
query vocab.      & 119,069 & 30,844 & 31,028 & 134,397 \\ \midrule
tokens\,/\,passage   & 169.5                        & 168.6                       & 168.1                       & 169.3                       \\
entities\,/\,passage & 17.8                         & 17.5                        & 17.3                        & 17.8                        \\
tokens\,/\,query  & 21.3                         & 22.1                        & 22.2                        & 21.4                        \\ \bottomrule
\end{tabular}
\caption{Statistics of \ReCoRD}
\label{tab:statistics}
\end{table}

\begin{table*}[!t]
\centering
\small
\begin{tabular}{@{}m{2cm}m{4.8cm}m{7cm}r@{}}
\toprule
\multicolumn{1}{c}{Reasoning} & \multicolumn{1}{c}{Description}                                                                                                                                                                                                             & \multicolumn{1}{c}{Example} & \multicolumn{1}{r}{\%} \\ \midrule
Paraphrasing                  & The answer sentence can be found by paraphrasing the query with some syntactic or lexical variation.                                                                                                               &    
\noindent\textbf{P:} \ldots\underline{\textcolor{blue}{Ralph Roberts}}\dots  then acquired other cable systems, changed the name of the company to \underline{\textcolor{blue}{Comcast}} and ran the company until he was aged 82

\noindent\textbf{Q:} \textcolor{red}{$\mathbf{X}$} began acquiring smaller cable systems and built the company into the nation's fifth-largest by 1988.

\noindent\textbf{A:} [Ralph Roberts]
& 3\%                            \\ \midrule
Partial Clue                  & Although a complete semantic match cannot be found between the query and the passage, the answer can be inferred through partial clues, such as some word/concept overlap. &     
\noindent\textbf{P:}\ldots
\underline{\textcolor{blue}{Hani Al}}-\underline{\textcolor{blue}{Sibai}} says he has `severe mobility problems' to get disability cash\ldots

\noindent\textbf{Q:} However the photographs caught \textcolor{red}{$\mathbf{X}$}-Sibai walking with apparent ease in the sunshine.

\noindent\textbf{A:} [Hani Al]
& 10\%                           \\ \midrule
Multi-sentence Reasoning      & It requires anaphora, or higher-level fusion of multiple sentences to find the answer.                                                                                                                                                      &      
\noindent\textbf{P:} \underline{\textcolor{blue}{Donald Trump}} is officially a \$10 billion man\ldots HIs campaign won't release a copy of the financial disclosure even though the \underline{\textcolor{blue}{FEC}} says it can do so on its own\ldots

\noindent\textbf{Q:} The \textcolor{red}{$\mathbf{X}$} campaign did provide a one-page summary of the billionaire's investment portfolio, which is remarkably modest for a man of his means.

\noindent\textbf{A:} [Donald Trump]
& 6\%                            \\ \midrule
Commonsense Reasoning        & It requires inference drew on common sense as well as multi-sentence reasoning to find the answer.                                                                                                &         
\noindent\textbf{P:} 
\ldots\underline{\textcolor{blue}{Daniela Hantuchova}} knocks \underline{\textcolor{blue}{Venus Williams}} out of \underline{\textcolor{blue}{Eastbourne}} 6-2 5-7 6-2 \ldots

\noindent\textbf{Q:} Hantuchova breezed through the first set in just under 40 minutes after breaking Williams' serve twice to take it 6-2 and led the second 4-2 before \textcolor{red}{$\mathbf{X}$} hit her stride.

\noindent\textbf{A:} [Venus Williams]
& 75\%                           \\ \midrule
Ambiguous                     & The passage is not informative enough, or the query does not have a unique answer.                                                                                                                                                       &         
\noindent\textbf{P:} The supermarket wars have heated up with the chief executive of \underline{\textcolor{blue}{Wesfarmers}} suggesting successful rival \underline{\textcolor{blue}{Aldi}} may not be paying its fair share of tax in \underline{\textcolor{blue}{Australia}}\ldots

\noindent\textbf{Q:} \textcolor{red}{$\mathbf{X}$}'s average corporate tax rate for the last three years was almost 31 per cent of net profit, and in 2013 it paid \$81.6 million in income tax.

\noindent\textbf{A:} [Aldi]
& 6\%                            \\ \bottomrule
\end{tabular}
\caption{An analysis of types of reasoning needed in 100 random samples from the dev. set of \ReCoRD.}
\label{tab:reason-types}
\end{table*}

\section{Data Analysis}
\label{sec:data-analysis}
\ReCoRD differs from other reading comprehension datasets due to its unique requirement for reasoning more than just paraphrasing.
In this section, we provide a qualitative analysis of \ReCoRD which highlights its unique features.

\noindent\textbf{Reasoning Types} We sampled 100 examples from the development set, and then manually categorized them into types shown in \cref{tab:reason-types}.
The results show that significantly different from existing datasets such as SQuAD~\cite{squad}, and NewsQA~\cite{newsqa}, \ReCoRD requires commonsense reasoning to answer 75\% of queries.
Owing to the machine filtering stage, only 3\% queries could be answered by paraphrasing.  
The small percentage (6\%) of ambiguous queries demonstrate the benefit of the human filtering stage.
We also noticed that 10\% queries can be answered through partial clues.
As the example shows, some of partial clues were caused by the incompleteness of named entity recognition in the stage of news article curation.

\noindent\textbf{Types of Commonsense Reasoning} 
Formalizing the commonsense knowledge needed for even simple reasoning problems is a huge undertaking.
Based on the observation of the sampled queries that required commonsense reasoning, we roughly categorized them into the following four coarse-gained types:

\begin{itemize}[itemsep=1pt,topsep=1pt,leftmargin=8pt]
    \item[]\textbf{Conceptual Knowledge}: the presumed knowledge of properties of concepts~\cite{wordnet,conceptnet,class-attributes,joci}.
    \item[]\textbf{Causal Reasoning}: the causal bridging inference invoked between two events, which is validated against common sense~\cite{bridging-inference,copa}. 
    \item[]\textbf{Na\"ive Psychology}: the predictable human mental states in reaction to events~\cite{naive-psychology}.
    \item[]\textbf{Other}: Other types of common sense, such as social norms, planning, spatial reasoning, etc.
\end{itemize}

We annotated one or more types to each of these queries, and computed the percentage of them in these queries as shown in \Cref{tab:commonsense-types}.

\begin{table*}[!t]
\centering
\small
\begin{tabular}{@{}m{2cm}m{12cm}r@{}}
\toprule
\multicolumn{1}{l}{Reasoning} & \multicolumn{1}{c}{Example} & \multicolumn{1}{r}{\%} \\ \midrule
Conceptual Knowledge  &    
\noindent\textbf{P:} Suspended hundreds of feet in the air amid glistening pillars of ice illuminated with ghostly lights from below, this could easily be a computer-generated scene from the latest sci-fi blockbuster movie. But in fact these ethereal photographs were taken in real life\ldots
captured by photographer \underline{\textcolor{blue}{Thomas Senf}} as climber \underline{\textcolor{blue}{Stephan Siegrist}}, 43, scaled frozen waterfall\ldots

\noindent\textbf{Q:} With bright lights illuminating his efforts from below, Mr \textcolor{red}{$\mathbf{X}$} appears to be on the set of a sci-fi movie.

\noindent\textbf{A:} [Stephan Siegrist]

\noindent\textbf{Commonsense knowledge:} \emph{Scenes such as ``a person suspended hundreds of feet in the air amid glistening pillars of ice illuminated with ghostly lights from below'' tend to be found in sci-fi movies.}
& 49.3\%                            \\ \midrule
Causal 

Reasoning   &     
\noindent\textbf{P:} \ldots
\underline{\textcolor{blue}{Jamie Lee Sharp}}, 25, stole keys to \pounds 40,000 \underline{\textcolor{blue}{Porsche Boxster}} during raid\ldots
He filmed himself boasting about the car before getting behind the wheel

\noindent\textbf{Q:} \textcolor{red}{$\mathbf{X}$} was  jailed for four years after pleading guilty to burglary, aggravated vehicle taking, driving whilst disqualified, drink-driving and driving without insurance.

\noindent\textbf{A:} [Jamie Lee Sharp]

\noindent\textbf{Commonsense knowledge:} \emph{If a person steals a car, the person may be arrested and jailed.}
& 32.0\%                           \\ \midrule
Na\"ive 

Psychology &      
\noindent\textbf{P:} \underline{\textcolor{blue}{Uruguay}} star \underline{\textcolor{blue}{Diego Forlan}} said Monday that he is leaving \underline{\textcolor{blue}{Atletico Madrid}} and is set to join \underline{\textcolor{blue}{Serie A}} \underline{\textcolor{blue}{Inter Milan}}\ldots \underline{\textcolor{blue}{Forlan}} said ``\ldots At the age of 33, going to a club like \underline{\textcolor{blue}{Inter}} is not an opportunity that comes up often\ldots"

\noindent\textbf{Q:} ``I am happy with the decision that I have taken, it is normal that some players come and others go," \textcolor{red}{$\mathbf{X}$} added.

\noindent\textbf{A:} [Diego Forlan, Forlan]

\noindent\textbf{Commonsense knowledge:} \emph{If a person has seized an valuable opportunity, the person will feel happy for it.}
& 28.0\%                            \\ \midrule
Other &         
\noindent\textbf{P:} 
A \underline{\textcolor{blue}{British}} backpacker who wrote a romantic note to locate a handsome stranger after spotting him on a \underline{\textcolor{blue}{New Zealand}} beach has finally met her \underline{\textcolor{blue}{Romeo}} for the first time. \underline{\textcolor{blue}{Sarah Milne}}, from \underline{\textcolor{blue}{Glasgow}}, left a handmade poster for the man, who she saw in \underline{\textcolor{blue}{Picton}} on Friday\ldots
She said she would return to the same spot in \underline{\textcolor{blue}{Picton}}, \underline{\textcolor{blue}{New Zealand}}, on Tuesday in search for him\ldots
 \underline{\textcolor{blue}{William Scott Chalmers}} revealed himself as the man and went to meet her\ldots

\noindent\textbf{Q:} Mr Chalmers, who brought a bottle of champagne with him, walked over to where Milne was sitting and said ``Hello, I'm \textcolor{red}{$\mathbf{X}$}, you know you could have just asked for my number.''

\noindent\textbf{A:} [William Scott Chalmers]

\noindent\textbf{Commonsense knowledge:} \emph{When two people meet each other for the first time, they will likely first introduce themselves.}
& 12.0\%                            \\ \bottomrule
\end{tabular}
\caption{An analysis of specific types of commonsense reasoning in 75 random sampled queries illustrated in \Cref{tab:reason-types} which requires common sense reasoning. A query may require multiple types of commonsense reasoning.}
\label{tab:commonsense-types}.
\end{table*}

\section{Evaluation}
\label{sec:evaluation}

We are interested in the performance of existing MRC architectures on \ReCoRD.
According to the task definition in~\Cref{sec:task}, \ReCoRD can be formalized as two types of machine reading comprehension (MRC) datasets: passages with cloze-style queries, or passages with queries whose answers are spans in the passage.
Therefore, we can evaluate two types of MRC models on \ReCoRD, and compare them with human performance.
All the evaluation is carried out based on the train\,/dev.\,/test split as illustrated in \Cref{tab:statistics}.

\subsection{Methods}

\noindent\textbf{DocQA}\footnote{\url{https://github.com/allenai/document-qa}}~\cite{docqa} is a strong baseline model for queries with extractive answers.
It consists of components such as bi-directional attention flow~\cite{bidaf} and self attention which are widely used in MRC models.
We also evaluate DocQA with ELMo~\cite{elmo} to analyze the impact of largely pre-trained encoder on our dataset.

\noindent\textbf{QANet}\footnote{The official implementation of QANet is not released. We use the implementation at \url{https://github.com/NLPLearn/QANet}.}~\cite{qanet} is one of the top MRC models for SQuAD-style datasets.
It is different from many other MRC models due to the use of transformer~\cite{transformer}.
Through QANet, we can evaluate the reasoning ability of transformer on our dataset.

\noindent\textbf{SAN}\footnote{\url{https://github.com/kevinduh/san_mrc}}~\cite{san} is also a top-rank MRC model.
It shares many components with DocQA, and employs a stochastic answer module.
Since we used SAN to filter out easy queries in our data collection, it is necessary to verify that the queries we collect is hard for not only SAN but also other MRC architectures.

\noindent\textbf{ASReader}\footnote{\url{https://github.com/rkadlec/asreader}}~\cite{asreader} is a strong baseline model for cloze-style datasets such as~\cite{deepmind-cnn-dailymail,cbt}.
Unlike other baseline models which search among all text spans in the passage, ASReader directly predicts answers from the candidate named entities. 

\noindent\textbf{Language Models}\footnote{\url{https://github.com/tensorflow/models/tree/master/research/lm_commonsense}} (LMs)~\cite{google-lms} trained on large corpora recently achieved the state-of-the-art scores on the Winograd Schema Challenge~\cite{wsc}.
Following in the same manner, we first concatenate the passage and the query together as a long sequence, and substitute $\mathbf{X}$ in the long sequence with each candidate entity; we use LMs to compute the probability of each resultant sequence and the substitution that results in the most probable sequence will be the predicted answer.

\noindent\textbf{Random Guess} acts as the lower bound of the evaluated models. It considers the queries in our dataset as cloze style, and randomly picks a candidate entity from the passage as the answer.

\subsection{Human Performance}
As described in \Cref{sec:human-filtering}, we obtained two worker answers for each query in the development and test sets, and confirmed that each query has been correctly answered by two different workers. 
To get human performance, we obtained an additional worker answer for each query, and compare it with the reference answers.

\subsection{Metrics}
We use two evaluation metrics similar to those used by SQuAD~\cite{squad}.
Both ignore punctuations and articles (e.g., \emph{a, an, the}).

\noindent\textbf{Exact Match} (EM) measures the percentage of predictions that match any one of the reference answers exactly.

\noindent(Macro-averaged) \textbf{F1} measures the average overlap between the prediction and the reference answers. 
We treat the prediction and the reference answer as bags of tokens, and compute their F1.
We take the maximum F1 over all of the reference answers for a given query, and then average over all of the queries.

\subsection{Results}
We show the evaluation results in \Cref{tab:performance}.
Humans are able to get 91.31 EM and 91.69 F1 on the set, with similar results on the development set.
In contrast, the best automatic method -- DocQA with ELMo --  achieves 45.44 EM and 46.65 F1 on the test set, illustrating a significant gap between human and machine reading comprehension on \ReCoRD.
All other methods without ELMo get EM/F1 scores significantly lower than DocQA with ELMo,
which shows the positive impact of ELMo (see in \Cref{sec:result-analysis}).
We also note that SAN leads to a result comparable with other strong baseline methods. 
This confirms that since SAN shares general components with many MRC models, using it to do machine filtering does help us filter out queries that are relatively easy to all the methods we evaluate. 
Finally, to our surprise, the unsupervised method (i.e., LM) which achieved the state-of-the-art scores on the Winograd Schema Challenge only leads to a result similar to the random guess baseline: a potential explanation is the lack of domain knowledge on our dataset.
We leave this question for future work.

\begin{table}[!t]
\centering
\small
\begin{tabular}{@{}lrr|rr@{}}
\toprule
\multirow{2}{*}{}                 & \multicolumn{2}{c}{Exact Match}                      & \multicolumn{2}{|c}{F1}                             \\ \cmidrule(l){2-5} 
                                  & \multicolumn{1}{c}{Dev.} & \multicolumn{1}{c|}{Test} & \multicolumn{1}{c}{Dev.} & \multicolumn{1}{c}{Test} \\ \midrule
\multicolumn{1}{l|}{Human}        & \textbf{91.28}           & \textbf{91.31}            & \textbf{91.64}           & \textbf{91.69}           \\ \midrule\midrule
\multicolumn{1}{l|}{DocQA w/ ELMo}        & 44.13                    & 45.44                     & 45.39                    & 46.65                    \\ 
\multicolumn{1}{l|}{DocQA w/o ELMo}  & 36.59                    & 38.52                     & 37.89                    & 39.76                    \\ \midrule
\multicolumn{1}{l|}{SAN}          & 38.14                    & 39.77                     & 39.09                    & 40.72                    \\ \midrule
\multicolumn{1}{l|}{QANet}        & 35.38                    & 36.51                     & 36.75                    & 37.79                    \\ \midrule\midrule
\multicolumn{1}{l|}{ASReader}     & 29.24                    & 29.80                     & 29.80                    & 30.35                    \\ \midrule
\multicolumn{1}{l|}{LM}           & 16.73                    & 17.57                     & 17.41                    & 18.15                    \\ \midrule
\midrule
\multicolumn{1}{l|}{Random Guess} & 18.41                    & 18.55                     & 19.06                    & 19.12                    \\ \bottomrule
\end{tabular}
\caption{Performance of various methods and human.}
\label{tab:performance}
\end{table}
\kev{Are you shrinking Table 4 too much? Make sure you don't violate submission rules}

\subsection{Analysis}
\label{sec:result-analysis}


\noindent\textbf{Human Errors}
About 8\% dev./test queries have not been correctly answered in the human evaluation.
We analyzed samples from these queries, and found that in most queries human was able to narrow down the set of possible candidate entities, but not able to find a unique answer. 
In many cases, two candidate entities equally fit $\mathbf{X}$ unless human has the specific background knowledge. 
We show an example in the~\Cref{sec:case-study}.

For the method analysis, we mainly analyzed the results of three representative methods: DocQA w/ ELMo, DocQA, and QANet.

\begin{figure}[!ht]
\centering
\includegraphics[width=0.40\textwidth]{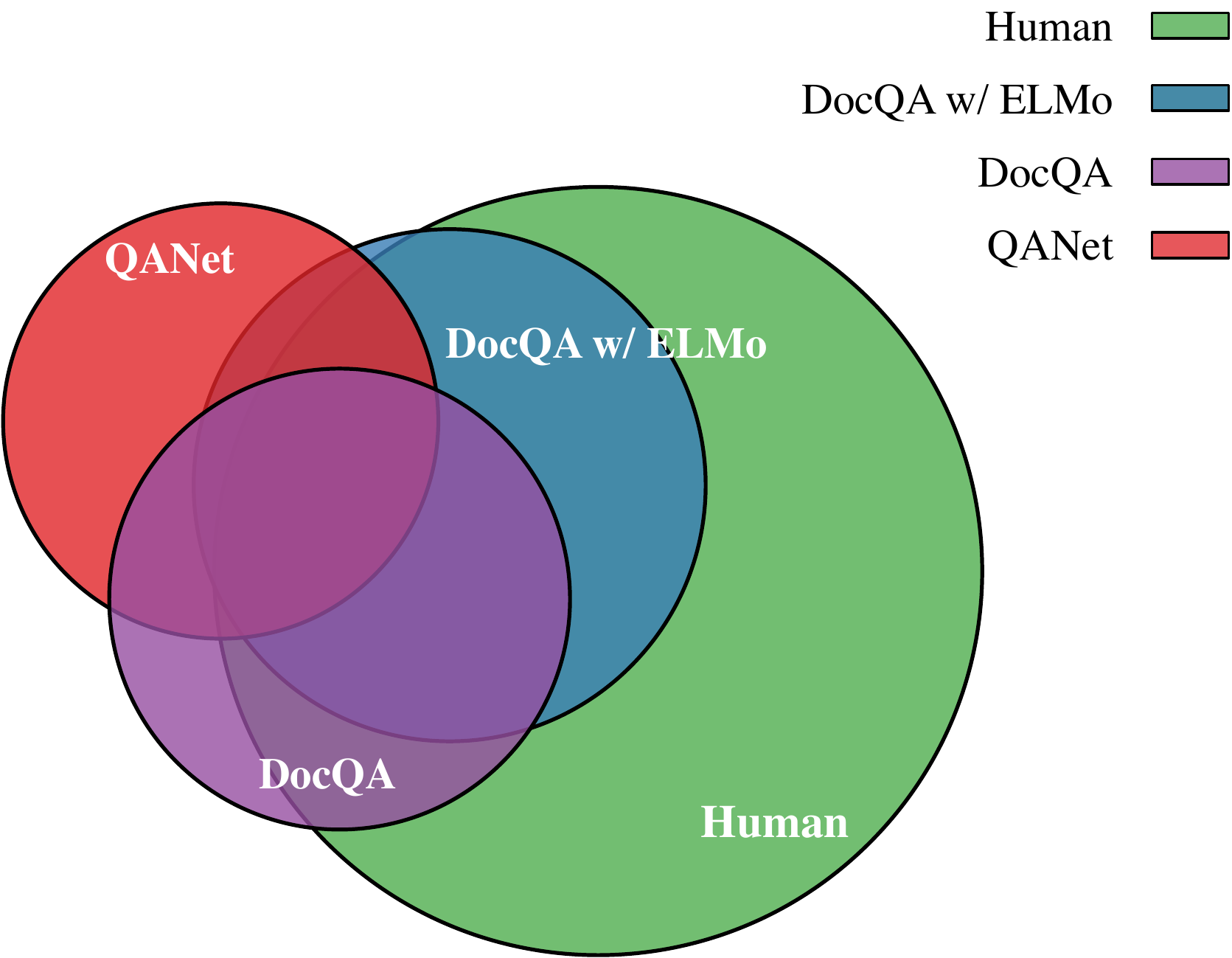}
\caption{The Venn diagram of correct predictions from various methods and human on the development set.\label{fig:venn}}
\end{figure}

\noindent\textbf{Impact of ELMo}
As shown in \Cref{fig:venn}, among all three methods the correct predictions of DocQA w/ ELMo have the largest overlap
(92.6\%) with the human predictions.
As an ablation study, we analyzed queries which were only correctly answered after ELMo was added.
We found that in some cases ELMo helped the prediction by incorporating the knowledge of language models.
We show an example in the~\Cref{sec:case-study}.

\noindent\textbf{Predictions of QANet}
\Cref{fig:venn} shows that QANet correctly answered some ambiguous queries, which we think was due to the randomness of parameter initialization and did not reflect the true reasoning ability. 
Since QANet uses the transformer-based encoder and DocQA uses the LSTM-based encoder, 
we see a significant difference of predictions between QANet and DocQA.

\begin{table}[!ht]
\centering
\begin{tabular}{@{}lr@{}}
\toprule
\multicolumn{1}{c}{Method} & \multicolumn{1}{c}{OOC Rate} \\ \midrule
DocQA w/ ELMo              & 6.27\%                                \\
DocQA                      & 6.37\%                                \\
QANet                      & 6.41\%                                \\ \bottomrule
\end{tabular}
\caption{The out-of-candidate-entities (OOC) rate of three analyzed methods.}
\label{tab:ooc}
\end{table}

\noindent\textbf{Impact of Cloze-style Setting}
Except ASReader, all the MRC models were evaluated under the extractive setting, which means the information of candidate named entities was not used.
Instead, extractive models searched answers from all possible text spans in passages.
To show the potential benefit of using the candidate entities in these models, we computed the percentage of model predictions that could not be found in the candidate entities. 
As shown in \Cref{tab:ooc}, all three methods have about 6\% OOC predictions. 
Making use of the candidate entities would potentially help them increase the performance by 6\%.

\begin{figure}[!ht]
\centering
\includegraphics[width=0.49\textwidth]{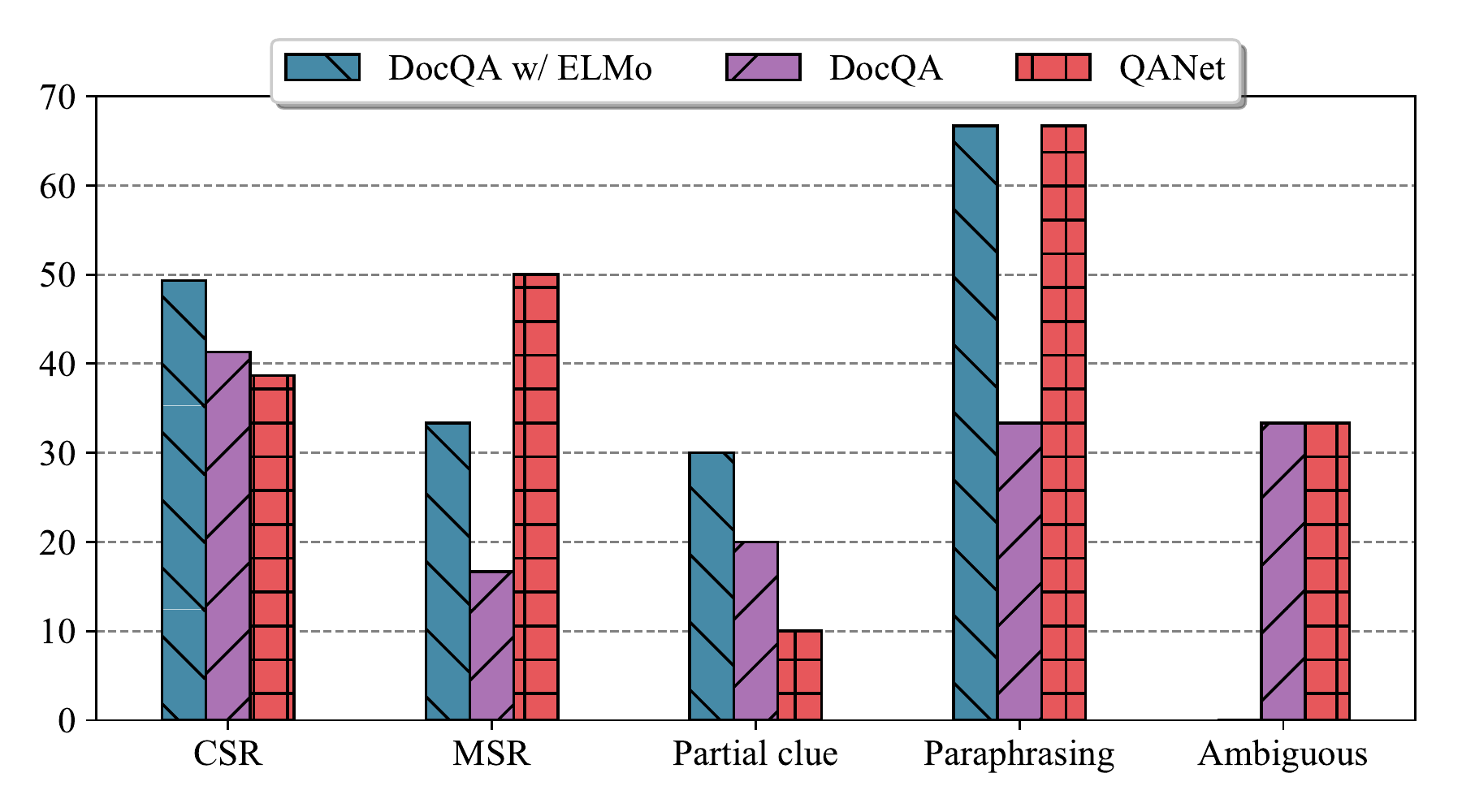}
\caption{Performance of three analyzed methods on the 100 random samples with reasoning types labeled.(CSR stands for commonsense reasoning, and MSR stands for multi-sentence reasoning.)\label{fig:baselines-eval-on-question-types}}
\end{figure}

\begin{figure}[!ht]
\centering
\includegraphics[width=0.49\textwidth]{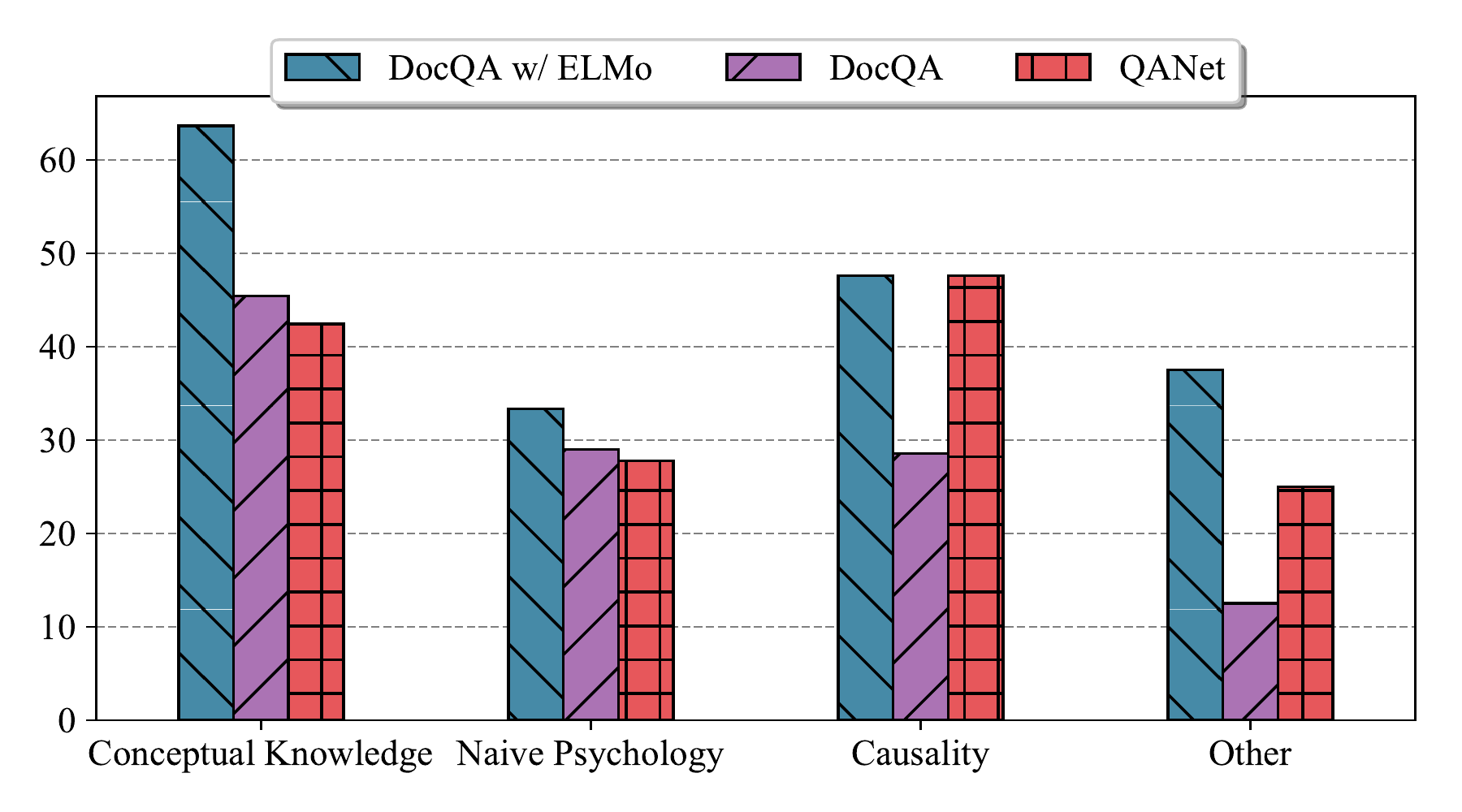}
\caption{Performance of three analyzed methods on 75\% of the random samples with specific commonsense reasoning types labeled.\label{fig:baselines-eval-on-commonsense-types}}
\end{figure}

In \Cref{sec:data-analysis}, we manually labeled 100 randomly sampled queries with different types of reasoning.
In \Cref{fig:baselines-eval-on-question-types} and \ref{fig:baselines-eval-on-commonsense-types}, we show the performance of three analyzed methods on these queries. 

\Cref{fig:baselines-eval-on-question-types} shows that three methods performed poorly on queries requiring commonsense reasoning, multi-sentence reasoning and partial clue.
Compared to DocQA, QANet performed better on multi-sentence reasoning queries probably due to the use of transformer.
Also, QANet outperformed DocQA on paraphrased queries probably because we used SAN to filtering queries and SAN has an architecture similar to DocQA.
As we expect, ELMo improved the performance of DocQA on paraphrased queries.

Among the 75\% sampled queries that require commonsense reasoning, we see that ELMo significantly improved the performance of commonsense reasoning with presumed knowledge.
For all other types of commonsense reasoning, all three methods have relatively poor performance.

\section{Related Datasets}
\label{sec:related-datasets}

\ReCoRD relates to two strands of research in datasets: data for reading comprehension, and that for commonsense reasoning.

\noindent\textbf{Reading Comprehension}
\emph{The CNN/Daily Mail Corpus} \cite{deepmind-cnn-dailymail}, \emph{The Children's Book Test} (CBT)~\cite{cbt}, and LAMBADA~\cite{lambada} are closely related to \ReCoRD:
(1) \emph{The CNN/Daily Mail Corpus} constructed queries from the bullet points, most of which required limited reasoning ability~\cite{exam-of-cnn-dailymail}.
(2) CBT is a collection of 21 consecutive sentences from book excerpts, with one
word randomly removed from the last sentence.
Since CBT has no machine or human filtering to ensure quality,
only a small portion of the CBT examples really probes machines' ability to understand the context.
(3) Built in a similar manner to CBT, LAMBADA was filtered to be human-guessable in the broader context only.
Differing from \ReCoRD, LAMBADA was designed to be a language modeling problem where contexts were not required to be event summaries, and answers were not necessarily in the context.

Since all candidate answers were extracted from in the passage, \ReCoRD can also be formalized as a extractive MRC dataset,
similar to SQuAD~\cite{squad} and NewsQA~\cite{newsqa}.
The difference is that questions in these datasets were curated from crowdworkers.
Since it is hard to control the quality of crowdsourced questions, a large portion of questions in these datasets can be answered by word matching or paraphrasing~\cite{adversarial-squad,squad-v2,adversarial-training}. 
There are other large-scale datasets~\cite{msmarco,triviaqa,race,searchqa,narrativeqa,coqa,quac,hotpotqa} targeting different aspects of reading comprehension. See \cite{gaosurvey} for a recent survey.

\noindent\textbf{Commonsense Reasoning}
ROCStories Corpus~\cite{rocstories}, SWAG~\cite{swag}, and \emph{The Winograd Schema Challenge}~(WSC) \cite{wsc} are related \ReCoRD:
(1) ROCStories assesses commonsense reasoning in story understanding by choosing the correct story ending from only two candidates. Stories in the corpus were all curated from crowdworkers, which could suffer from human elicitation bias~\cite{reporting-bias,vision-reporting-bias,joci}.
(2) SWAG unifies commonsense reasoning and natural language inference. 
It selects an ending from multiple choices which is most likely to be anticipated from the situation describe in the premise.
The counterfactual endings in SWAG were generated using language models with adversarial filtering.
(3) WSC foucses on intra-sentential pronoun disambiguation problems that require commonsense reasoning.
There are other datasets~\cite{copa,joci,naive-psychology-in-stories,event2mind} targeting different aspects of commonsense reasoning.

\section{Conclusion}
We introduced \ReCoRD, a large-scale reading comprehension dataset requiring commonsense reasoning.
Unlike existing machine reading comprehension (MRC) datasets, \ReCoRD contains a large portion of queries that require commonsense reasoning to be answered.
Our baselines, including top performers on existing MRC datasets, are no match for human competence on \ReCoRD.
We hope that \ReCoRD will spur more research in MRC with commonsense reasoning.

\bibliography{naaclhlt2019}
\bibliographystyle{acl_natbib}

\clearpage

\appendix

\section{Appendices}
\label{sec:appendix}

\subsection{Case Study}
\label{sec:case-study}

\noindent\textbf{Human Error}
\Cref{tab:human-error-ex} shows an example where the ambiguous query caused human error.
The passage in this example describes ``\emph{ambiverts}'', and there are two experts studying it:
``\emph{Vanessa Van Edwards}'' and ``\emph{Adam Grant}''.
Both of them fit in the query asking who gave advice to ambiverts.
There is no further information to help human choose a unique answer for this query.

\begin{table}[!ht]
\centering
\small
\begin{tabular}{m{0.45\textwidth}}
\toprule
\noindent\textbf{Passage:} Your colleagues think you're quiet, but your friends think you're a party animal. If that sounds like you, then you may be what psychologists describe as an 'ambivert'. Scientists believe around two-thirds of people are ambiverts; a personality category that has, up until now, been given relatively little attention. 'Most people who are ambiverts have been told the wrong category their whole life,' \underline{\textcolor{blue}{Vanessa Van Edwards}}, an \underline{\textcolor{blue}{Orgeon}}-based behavioural expert, told DailyMail.com 'You hear extrovert and you hear introvert, and you think 'ugh, that's not me'.' \underline{\textcolor{blue}{Ambiversion}} is a label that has been around for some time, but gained popularity in 2013 with a paper in the journal \underline{\textcolor{blue}{Psychological Science}}, by \underline{\textcolor{blue}{Adam Grant}} the \underline{\textcolor{blue}{University of Pennsylvania}}.
\begin{itemize}[itemsep=0pt,topsep=3pt,leftmargin=8pt]
	\item Most ambiverts have been labelled incorrectly their whole life
	\item They slide up and down personality spectrum depending on the situation
	\item \underline{\textcolor{blue}{Ambiverts}} are good at gaining people's trust and making their point heard
	\item They often feel pressure to mirror personality of the person they are with
\end{itemize}\\
\noindent\textbf{Query:} 'Read each situation more carefully,' \textcolor{red}{$\mathbf{X}$} advised ambiverts, 'and ask yourself, 'What do I need to do right now to be most happy or successful?''\\
\noindent\textbf{Reference answers:} Adam Grant\\
\bottomrule
\end{tabular}
\caption{An example illustrating a ambiguous query.}
\label{tab:human-error-ex}
\end{table}

\noindent\textbf{Impact of ELMo}
\Cref{tab:elmo-impact-ex} shows an example where DocQA w/ ELMo correctly answered but DocQA failed.
The passage in this example describes a woman artist ``\emph{Sarah Milne}'' who launched a public appeal to find a handsome stranger ``\emph{William Scott Chalmers}'', and invited him to meet her. 
The query asks the missing information in the greetings from ``\emph{William Scott Chalmers}'' when he went to meet ``\emph{Sarah Milne}''.
Our common sense about social norms tells us when two people meet each other for the first time, they are very likely to first introduce themselves.
In the query of this example, when Mr. Chalmers said ``\emph{Hello, I'm \ldots}'', it is very likely that he was introducing himself.
Therefore, the name of Mr Chalmer fit $\mathbf{X}$ best.

In this example, the prediction of DocQA without ELMo is ``\emph{New Zealand}'' which is not even close to the reference answer.
The benefit of using ELMo in this example is that its language model will help exclude ``\emph{New Zealand}'' from the likely candidate answers,
because ``\emph{I'm \ldots}'' is usually followed by a person name  rather than a location name. 
Such a pattern learnt by ELMo is useful in narrowing down candidiate answers in \ReCoRD.

\begin{table}[!ht]
\centering
\small
\begin{tabular}{m{0.45\textwidth}}
\toprule
\noindent\textbf{Passage:} A \underline{\textcolor{blue}{British}} backpacker who wrote a romantic note to locate a handsome stranger after spotting him on a \underline{\textcolor{blue}{New Zealand}} beach has finally met her \underline{\textcolor{blue}{Romeo}} for the first time. \underline{\textcolor{blue}{Sarah Milne}}, from \underline{\textcolor{blue}{Glasgow}}, left a handmade poster for the man, who she saw in \underline{\textcolor{blue}{Picton}} on Friday and described as 'shirtless, wearing black shorts with stars tattooed on his torso and running with a curly, bouncy and blonde dog'. In her note, entitled 'Is this you? ', she invited the mystery stranger to meet her on the same beach  on Tuesday. But the message soon became a source of huge online interest with the identity of both the author and its intended target generating unexpected publicity.
\begin{itemize}[itemsep=0pt,topsep=3pt,leftmargin=8pt]
	\item \underline{\textcolor{blue}{Sarah Milne}}, a \underline{\textcolor{blue}{Glasgow}} artist, launched a public appeal to find the mystery man
	\item She wrote a heart-warming message and drew a picture of him with his dog
	\item She said she would return to the same spot in \underline{\textcolor{blue}{Picton}}, \underline{\textcolor{blue}{New Zealand}}, on Tuesday in search for him
	\item \underline{\textcolor{blue}{William Scott Chalmers}} revealed himself as the man and went to meet her
	\item He told \underline{\textcolor{blue}{Daily Mail Australia}} that he would ask her out for dinner
\end{itemize}\\
\noindent\textbf{Query:} Mr Chalmers, who brought a bottle of champagne with him, walked over to where Milne was sitting and said 'Hello, I'm \textcolor{red}{$\mathbf{X}$}, you know you could have just asked for my number.'\\
\noindent\textbf{Reference answers:} William Scott Chalmers\\
\bottomrule
\end{tabular}
\caption{An example illustrating the impact of ELMo.}
\label{tab:elmo-impact-ex}
\end{table}

\subsection{HIT Instructions}
\label{sec:hit-instructions}
\begin{figure*}[!ht]
\centering
\includegraphics[width=0.80\textwidth]{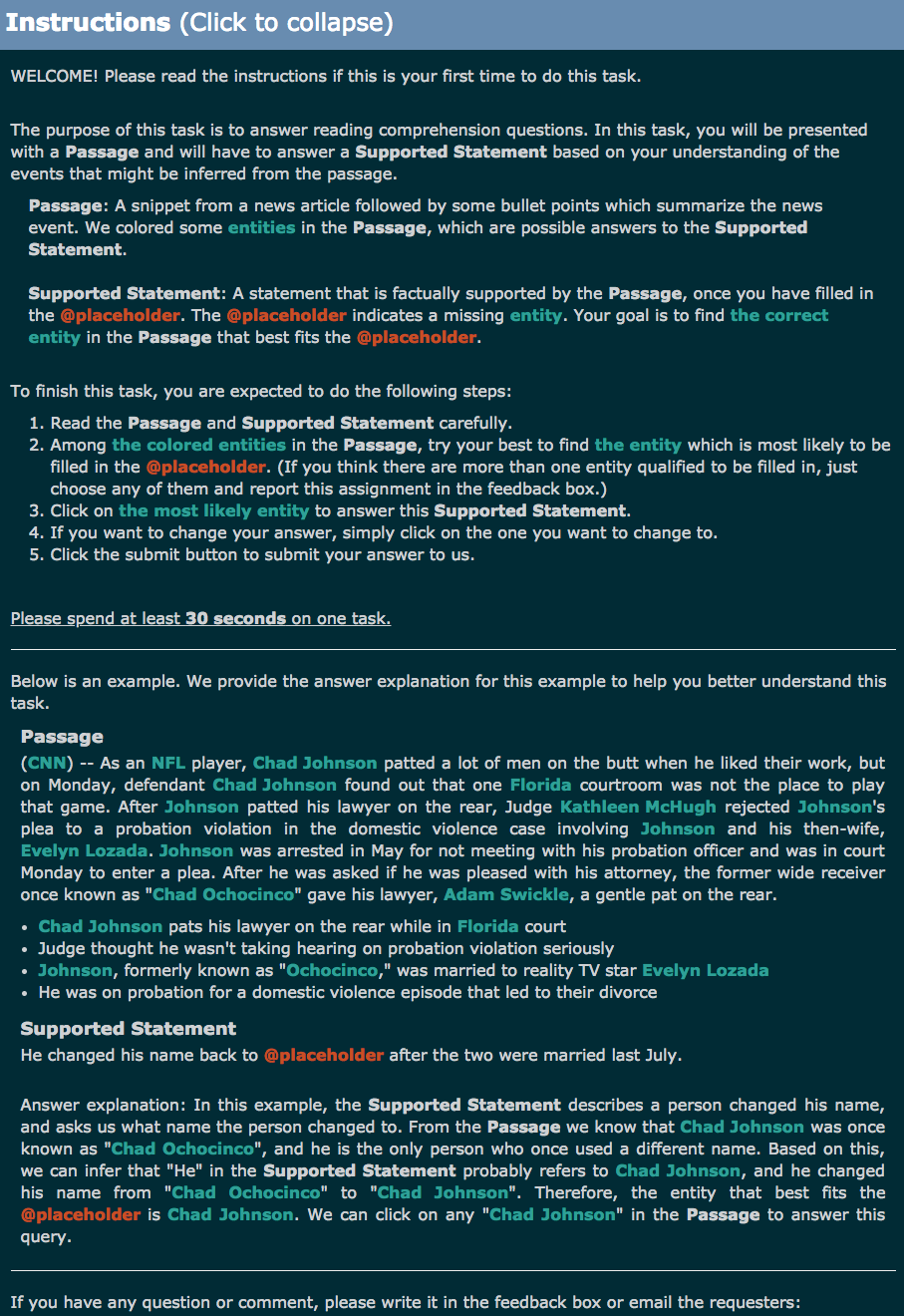}
\caption{Amazon Mechanical Turk HIT Instructions.\label{fig:hit-instructions}}
\end{figure*}

We show the instructions for Amazon Mechanical Turk HITs in \Cref{fig:hit-instructions}.

\end{document}